\documentclass[10pt,twocolumn,letterpaper]{article}

\usepackage{titling} %
\usepackage{cvpr}
\usepackage{times}
\usepackage{epsfig}
\usepackage{graphicx}
\usepackage{amsmath}
\usepackage{amssymb}

\usepackage{booktabs}
\usepackage{multirow}
\usepackage[dvipsnames,table]{xcolor}
\usepackage{enumitem}
\usepackage{dsfont}
\usepackage{balance}
\usepackage{placeins}

\usepackage[pagebackref=true,breaklinks=true,colorlinks,citecolor=Green,bookmarks=false]{hyperref}

\newcommand{\Cloudnetworks}{Permutation-equivariant networks}
\newcommand{\cloudnetworks}{permutation-equivariant networks}
\newcommand{\cloudnetwork}{permutation-equivariant network}
\newcommand{\normalize}{\eta}

\newcommand{\bmu}{{\boldsymbol{\mu}}}

\newcommand{\bb}{\mathbf{b}}

\newcommand{\bO}{\mathbf{O}}
\newcommand{\bw}{\mathbf{w}}
\newcommand{\bp}{\mathbf{p}}
\newcommand{\bP}{\mathbf{P}}

\newcommand{\bF}{\mathbf{F}}
\newcommand{\bbf}{\mathbf{f}}

\newcommand{\bsigma}{{\boldsymbol{\sigma}}}
\newcommand{\btheta}{{\boldsymbol{\theta}}}

\newcommand{\bW}{\mathbf{W}}
\newcommand{\bX}{\mathbf{X}}

\newcommand{\bv}{\mathbf{v}}
\newcommand{\bx}[0]{\mathbf{x}}

\newcommand{\by}[0]{\mathbf{y}}

\newcommand{\IR}{\mathds{R}}
\newcommand{\IE}{\mathds{E}}

\newcommand{\ba}{\mathbf{a}}

\newcommand{\tbl}[1]{Table~\ref{tbl:#1}}
\newcommand{\secref}[1]{Section~\ref{sec:#1}}
\newcommand{\refsec}[1]{Section~\ref{sec:#1}}

\DeclareMathOperator*{\argmin}{\text{argmin}}

\newcommand{\softmax}{\text{softmax}}

\newcommand{\loss}{\mathcal{L}}

\definecolor{color1}{RGB}{0,199,1}
\definecolor{color2}{RGB}{224,43,28}

\newcommand{\f}{\bbf}
\newcommand{\CN}{\mathcal{N}_{\textit{CN}}} %
\newcommand{\ACN}{\mathcal{N}_{\textit{ACN}}} %
\newcommand{\mean}{\mu}
\newcommand{\std}{\sigma}

\newcommand{\w}{\mathbf{w}}
\newcommand{\bomega}{{\boldsymbol{\omega}}}
\newcommand{\attnet}{\mathcal{W}_\bomega}

\usepackage{overpic}
\usepackage{microtype}

\usepackage{color}
\definecolor{turquoise}{cmyk}{0.65,0,0.1,0.3}
\definecolor{purple}{rgb}{0.65,0,0.65}
\definecolor{dark_green}{rgb}{0, 0.5, 0}
\definecolor{orange}{rgb}{0.8, 0.6, 0.2}
\definecolor{red}{rgb}{0.8, 0.2, 0.2}
\definecolor{darkred}{rgb}{0.6, 0.1, 0.05}
\definecolor{blueish}{rgb}{0.0, 0.3, .6}
\definecolor{light_gray}{rgb}{0.7, 0.7, .7}
\definecolor{pink}{rgb}{1, 0, 1}
\definecolor{greyblue}{rgb}{0.25, 0.25, 1}

\usepackage{comment}
\specialcomment{DRAFT}{\color{red}}{\color{black}}

\newcommand{\Fig}[1]{Fig.~\ref{fig:#1}}
\newcommand{\Figure}[1]{Fig.~\ref{fig:#1}}

\newcommand{\eq}[1]{(\ref{eq:#1})}

\newcommand{\Section}[1]{Section~\ref{sec:#1}}

\newcommand{\Table}[1]{Table~\ref{tbl:#1}}

\newcommand{\CIRCLE}[1]{\raisebox{.5pt}{\footnotesize \textcircled{\raisebox{-.6pt}{#1}}}}
\renewcommand{\CIRCLE}[1]{{(#1)}}

\usepackage{blindtext}

\renewcommand{\paragraph}[1]{\vspace{0.4em}\noindent\textbf{#1}}

\cvprfinalcopy %

\def\cvprPaperID{308} %

\ifcvprfinal\fi
\begin{document}

\title{ACNe: Attentive Context Normalization \\ for Robust Permutation-Equivariant Learning}

\author{Weiwei Sun$^1$ \quad Wei Jiang$^1$ \quad Eduard Trulls$^2$ \quad Andrea Tagliasacchi$^3$ \quad Kwang Moo Yi$^1$\\
  $^1$University of Victoria \quad
  $^2$Google Research, Zurich \quad
  $^3$Google Research, Toronto\\
  \texttt{\small \{weiweisun, jiangwei, kyi\}@uvic.ca} \quad
  \texttt{\small \{trulls, taglia\}@google.com}
}

\maketitle
\begin{abstract}
Many problems in computer vision require dealing with sparse, unordered data in the form of point clouds.
\Cloudnetworks{} have become a popular solution~--~they operate on individual data points with simple perceptrons and extract contextual information with global pooling. 
This can be achieved with a simple normalization of the feature maps, a global operation that is unaffected by the order.
In this paper, we propose \textit{Attentive Context Normalization}~(ACN), a simple yet effective technique to build \cloudnetworks{} robust to outliers.
Specifically, we show how to normalize the feature maps with weights that are estimated within the network, excluding outliers from this normalization.
We use this mechanism to leverage two types of attention: local and global -- by combining them, our method is able to find the essential data points in high-dimensional space to solve a given task. 
We demonstrate through extensive experiments that our
approach, which we call Attentive Context Networks (ACNe),
provides a significant leap in performance compared to the state-of-the-art on camera pose estimation, robust fitting, and point cloud classification under noise and outliers.
Source code: \href{https://github.com/vcg-uvic/acne}{https://github.com/vcg-uvic/acne}.
\end{abstract}

\section{Introduction}
\label{sec:intro}

Several problems in computer vision require processing sparse, \textit{unordered} collections of vectors $\mathcal{P}\!=\!\{ \mathbf{p}_n\!\in\!\mathbb{R}^D \}$, commonly called \textit{clouds}.
Examples include pixel locations ($D{=}2$), point clouds from depth sensors ($D{=}3$), and sparse correspondences across a pair of images ($D{=}4$).
The latter includes wide-baseline stereo, one of the fundamental problems in computer vision.
It lies at the core of Structure-from-Motion (SfM), which, in turn, is the building block of applications such as 3D reconstruction~\cite{Agarwal09}, image-based rendering~\cite{photosynth} and time-lapse smoothing~\cite{hyperlapse}.

Wide-baseline stereo has been traditionally solved by extracting small collections of discrete \textit{keypoints}~\cite{Lowe04} and finding {\em correspondences} among them with robust estimators~\cite{Fischler81}, a reliable approach used for well over two decades.
This has changed over the past few years, with the arrival of deep learning and an abundance of new dense~\cite{Zamir16,Ummenhofer17,Zhou17a} and sparse~\cite{Yi18a,Zheng18,Ranftl18,Zhao19,Kluger20} methods.
Here,
we focus on sparse methods, which have seen many recent developments made possible by the introduction of \textit{PointNets}~\cite{Qi17a,Qi17b} -- neural networks that rely on multi-layer perceptrons and global pooling to process unordered data in a \textit{permutation-equivariant} manner -- something which is not feasible with neither convolutional nor fully-connected layers.

Networks of this type -- hereafter referred to as \textit{\cloudnetworks{}} -- have pioneered the application of deep learning to point clouds.
The original PointNet relied on the concatenation of point-wise (context-agnostic) and global (point-agnostic) features to achieve permutation equivariance.
Yi~et~al.~\cite{Yi18a} proposed \textit{Context Normalization} (CN) as a simple, yet effective alternative to global feature pooling: all it requires is a non-parametric normalization of the feature maps to zero mean and unit variance.
Contrary to other normalization techniques utilized by neural networks~\cite{Ioffe15,Ba16,Ulyanov16,Wu18}, whose primary objective is to improve convergence, context normalization is used to generate contextual information while preserving permutation equivariance.
Despite its simplicity, it proved more effective than the PointNet approach on wide-baseline stereo, contributing to a relative increase in pose estimation accuracy of $50{-}100\%$; see~\cite[Fig. 5]{Yi18a}.

\begin{figure*}
\centering
\includegraphics[width=0.95\linewidth, trim = 0 0 0 15, clip]{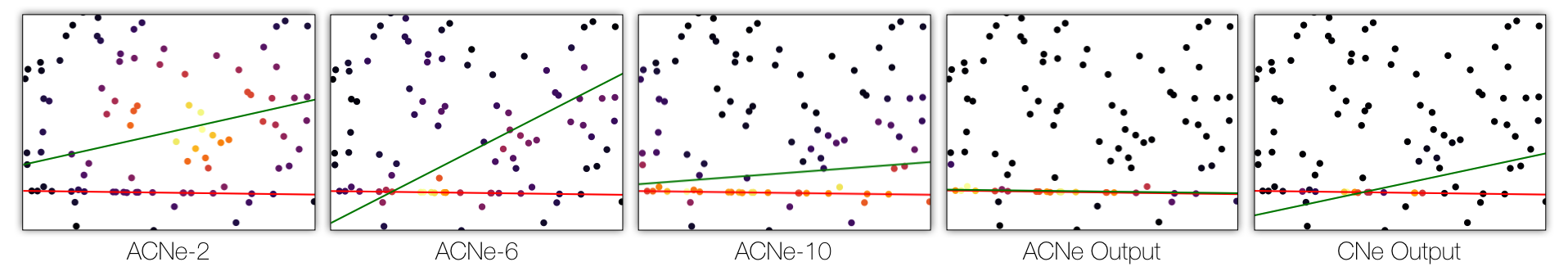}
\vspace{-0.4em}
\caption{
\textbf{Robust neural line fitting} --
We learn to fit lines with outliers (80\%) via our ACNe, as well as CNe~\cite{Yi18a}.
We visualize the {\bf \color{red} ground truth} and the {\bf{\color{dark_green} network estimates}}.
We color-code the weights learned by the k-th residual layer of ACNe and used to normalize the feature maps -- notice that our method, which mimics Iterative Re-weighted Least Squares~(IRLS), learns to {\em progressively} focus its attention on the inliers.
This allows ACNe to find the correct solution where CNe fails.
}
\vspace{-1em}
\label{fig:toy}
\end{figure*}

Note that CN normalizes the feature maps according to first- (mean) and second- (variance) order moments. 
Interestingly, these two quantities can be expressed as the solution of a least-squares problem:
\begin{align}
\hat{\bmu} &= \argmin_\bmu \sum_n \|\bp_n - \bmu \|_2^2
\label{eq:first}
\\
\hat{\bsigma} &= \argmin_\bsigma \sum_n \left\|\|\bp_n - \hat{\bmu} \|_2^2 - \bsigma^{\circ2}\right\|_2^2
\end{align}
However, it is well known that least-squares optimization is \textit{not robust} to outliers~\cite[Sec.~3]{regcourse_siga16}, a problem that also afflicts~CN.
We illustrate this limitation in \Figure{toy}, where the toy task is to fit a line to data corrupted by outliers.
Note that this is a {\em critical weakness}, as the application CN was originally devised for, wide-baseline stereo, a problem \textit{plagued with outliers} : outlier ratios above $80\%$ are typical in standard public datasets; see \Section{stereo}.

To address this issue, we take inspiration\footnote{See \Section{residual2weight} of the supplementary material. A robust kernel can also be trained with a neural network and enforcing monotonicity; see~\cite{Cotter19}.} from a classical technique used in robust optimization:~Iteratively Re-weighted Least Squares (IRLS)~\cite{Irls}.
As an example, let us consider the computation of the first-order moment~\eq{first}.
Rather than using the square of the residuals, we can optimize with respect to a \textit{robust} kernel $\kappa$ that allows for outliers to be ignored:
\vspace{-0.7em}
\begin{align}
\argmin_\bmu &\sum_n \kappa(\|\mathbf{p}_n - \bmu \|_2),
\label{eq:ls}
\end{align}
which can then be converted back into an \textit{iterative} least-squares optimization ($t$ indexes iterations):
\begin{align}
\argmin_{\bmu^t} &\sum_n \underbrace{{\psi(\|\mathbf{p}_n - \bmu^{t-1} \|_2)}^{-1}}_{\text{attention } w_n^t} \|\mathbf{p}_n - \bmu^t \|_2^2
,
\label{eq:irls}
\end{align}
where $\psi(\cdot)$ is the 
\textit{penalty function} associated with the kernel $\kappa(\cdot)$; see \cite{mestimator,robust}.
Inspired by this, we design a network that learns to \textit{progressively} focus its attention on the inliers, 
operating analogously to $\psi(.)$ over the IRLS iterations.

Specifically, we propose to train a perceptron that translates the (intermediate) feature maps into their corresponding attention weights, and normalizes them accordingly.
We denote this approach as {\em Attentive Context Normalization}~(ACN), and the networks that rely on this mechanism {\em Attentive Context Networks}~(ACNe).
We consider two types of attention, one that operates on each data point individually~(local), and one that estimates the relative importance of data points~(global), and demonstrate that using them together yields the best performance.
We also evaluate the effect of supervising this attention mechanism when possible. 
We verify the effectiveness of our method on \CIRCLE{1}~robust line fitting, \CIRCLE{2}~classification of 2D and 3D point clouds, and \CIRCLE{3}~wide-baseline stereo on real-world datasets (outdoors \textit{and} indoors) showing significant improvements over the state of the art.
Our work is, to the best our knowledge, the first to apply attentive mechanisms to the \emph{normalization} of feature maps.
One can also apply a more common form of attention by operating \textit{directly} on feature maps~\cite{Wang17b,Dusmanu19}, but we demonstrate that this does not perform as effectively.

\section{Related work}
\label{sec:related}

We discuss recent works on deep networks operating on point clouds, review various normalization methods for deep networks, and briefly discuss attention mechanisms.

\paragraph{Deep networks for point clouds.}
Several methods have been proposed to process point cloud data with neural networks. These include graph convolutional networks~\cite{Defferrard16,Kipf17},
VoxelNets~\cite{voxelnet}, tangent convolutions~\cite{Tatarchenko18}, and many others.
A simpler strategy was introduced by PointNets~\cite{Qi17a,Qi17b}, which has since become a popular solution due to its simplicity.
At their core, they are convolutional neural networks with $1 \times 1$ kernels and global pooling operations.
Enhancements to the PointNet architecture include incorporating locality information with kernel correlation~\cite{Shen18b}, and contextual information with LSTMs~\cite{Liu18f}.
Another relevant work is Deep Sets~\cite{Zaheer17}, which derives neural network parameterizations that guarantee permutation-equivariance.

\paragraph{\Cloudnetworks{} for stereo.}
While PointNets were originally introduced for segmentation and classification of 3D point clouds, Yi~et~al.~\cite{Yi18a} demonstrated that they can also be highly effective for robust matching in stereo, showing a drastic leap in performance against hand-crafted methods~\cite{Fischler81,Torr00,Bian17}.
The core ingredient of Yi~et~al.~\cite{Yi18a} is Context Normalization (CN), an alternative to global feature pooling from PointNets.
While similar to other normalization techniques for deep networks~\cite{Ioffe15,Ba16,Ulyanov16,Wu18}, CN has a different role -- to aggregate point-wise feature maps and generate \textit{contextual} information.
Follow-ups to CN include the use of architectures similar to Yi~et~al.~\cite{Yi18a} to iteratively estimate fundamental matrices~\cite{Ranftl18}, novel loss formulations~\cite{Zheng18}, and the modeling of locality~\cite{Zhao19}.
In~OANet~\cite{Zhang19a}, order-aware filtering was utilized to incorporate context and spatial correlation.
While all of these works rely on ``vanilla'' CN, we show how to improve its performance by embedding an attention mechanism therein.
Our improvements are compatible with \textit{any} of these techniques.

\paragraph{Normalization in deep networks.}
In addition to CN, different strategies have been proposed to normalize feature maps in a deep network, starting with the seminal work of Batch Normalization~\cite{Ioffe15}, which proposed to normalize the feature maps over a mini-batch.
Layer Normalization~\cite{Ba16} transposed this operation by looking at all channels for a single sample in the batch, whereas Group Normalization~\cite{Wu18} applied it over subsets of channels. Further efforts have proposed to normalize the weights instead of the activations~\cite{Salimans16}, or their eigenvalues~\cite{Miyato18}. The main use of all these normalization techniques is to stabilize the optimization process and speed up convergence. By contrast, Instance Normalization~\cite{Ulyanov16} proposed to normalize individual image samples for style transfer, and was improved upon in~\cite{Huang17} by aligning the mean and standard deviation of content and style.
Regardless of the specifics,
all of these normalization techniques operate on the entire sample -- in other words, they do not consider the presence of outliers or their statistics.
While this is not critical in image-based pipelines, it can be extremely harmful for point clouds; see
\Figure{toy}.

\paragraph{Attentional methods.}
The core idea behind attention mechanisms is to \textit{focus} on the crucial parts of the input. There are different forms of attention, and they have been applied to a wide range of machine learning problems, from natural language processing to images.
Vaswani~et~al.~\cite{Vaswani17} proposed an attentional model for machine translation eschewing recurrent architectures.
Luong~et~al.~\cite{Luong15} blended two forms of attention on sequential inputs, demonstrating performance improvements in text translation.
Xu~et~al.~\cite{Xu15} showed how to employ soft and hard attention to gaze on salient objects and generate automated image captions.
Local response normalization has been used to find salient responses in feature maps~\cite{Jarrett09,Krizhevsky12}, and can be interpreted as a form of lateral inhibition~\cite{Hartline56}.
The use of attention in convolutional deep networks was pioneered by Spatial Transformer Networks~\cite{Jaderberg15}, which introduced a differentiable sampler that allows for spatial manipulation of the image.
In~\cite{Xie18}, attention is directly applied to the feature map, given by a PointNet-style network operating on point clouds.
However, this strategy does not work as well as ours for wide-baseline stereo; see \Section{featmap_vs_norm} in the supplementary material.

\section{Attentive Context Normalization}
\label{sec:acn}
Given a feature map $\bbf{\in}\IR^{N\times C}$, where $N$ is the number of features~(or data points at layer zero), $C$ is the number of channels, and each row corresponds to a data point, we recall that Context Normalization~\cite{Yi18a} is a non-parametric operation that can be written as
\begin{equation}
\CN(\f) = (\f - \mean(\f)) \oslash \std(\f),
\end{equation}
where $\mean(\f){=}\IE[\f]$ is the arithmetic mean,  $\std(\f) {=} \sqrt{\IE[(\f - \IE[\f])^{\circ2}]}$ is the standard deviation of the features across $N$, and $\oslash$ denotes the element-wise division.
Here we assume a single cloud, but generalizing to multiple clouds (i.e.~batch) is straightforward.
Note that to preserve the properties of unstructured clouds, the information in the feature maps needs to be normalized in a \textit{permutation-equivariant} way.
We extend CN by introducing a weight vector $\w{\in}[0, \dots1]^{N}$, and indicate with $\mean_\w\left(\cdot\right)$ and $\std_\w\left(\cdot\right)$ the corresponding weighted mean and standard deviation.
In contrast to Context Normalization, we compute the weights $\w$ with a parametric function $\attnet(\cdot)$ with trainable parameters\footnote{For simplicity, we abuse the notation and drop the layer index from all parameters. All the perceptrons in our work operate individually over each data point with shared parameters across each layer.} $\bomega$ that takes as input the feature map, and returns a \textit{unit norm} vector of weights:
\begin{equation}
\w = \normalize\!\left(\attnet(\f)\right),
\quad 
\normalize(\bx)=\bx / \|\bx\|_1.
\label{eq:w}
\end{equation}
We then define \textit{Attentive Context Normalization} as
\begin{equation}
\ACN(\f ; \bw) = (\f - \mean_\w(\f)) \oslash \std_\w(\f).
\label{eq:acn}
\end{equation}
The purpose of the attention network $\attnet\left(\cdot\right)$ is to compute a weight function that focuses the normalization of the feature maps %
on a subset of the input features -- the inliers.
As a result, the network can learn to effectively {\em cluster} the features, and therefore separate inliers from outliers.

There are multiple attention functions that we can design, and multiple ways to combine them into a single attention vector $\w$.
We will now describe those that we found effective for finding correspondences in wide-baseline stereo, and how to combine and supervise them effectively.

\begin{figure*}
\centering
\includegraphics[width=.95\linewidth]{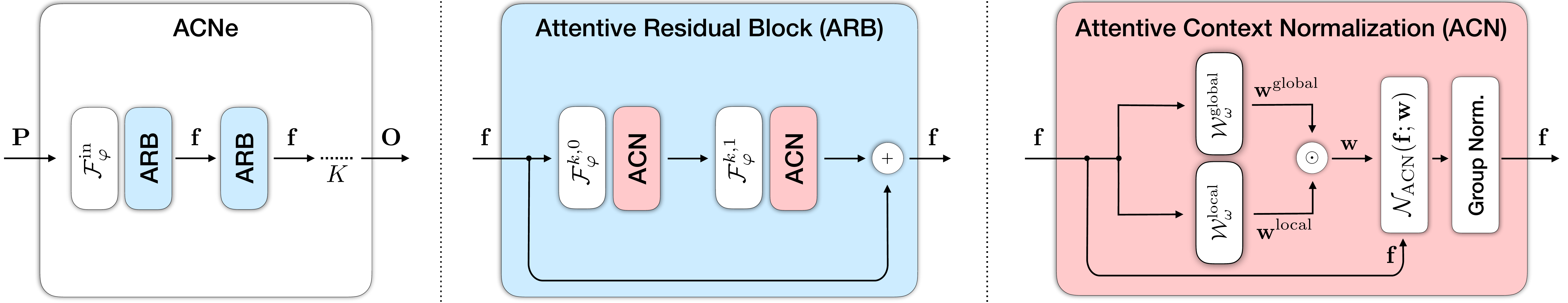}
\caption{
\textbf{ACNe architecture} -- 
(Left) Our \cloudnetwork{} receives an input tensor $\mathbf{P}$ of size $N {\times} D$, which is processed by a series of $K$ Attentive Residual Blocks (ARB).
The output of the network is a tensor $\bO$ size $N {\times} C$, which is then converted to a representation appropriate for the task at hand.
Note that the first perceptron $\mathcal{F}_\varphi^\text{in}$ changes the dimensionality from $\mathbf{P}$ of size $N {\times} D$ (input dimensions) to features $\mathbf{f}$ of size $N {\times} C$.
(Middle) Within each residual path of the ARB, we manipulate the feature map with perceptrons $\mathcal{F}_\varphi$ with parameters $\varphi$, followed by Attentive Context Normalization (ACN)~--~we repeat this structure twice.
(Right) An ACN module computes local/global attention with two trainable networks, combines them via element-wise multiplication, and normalizes the feature maps with said weights~--~the $\mathcal{N}_\text{ACN}$ block~--~followed by Group Normalization.
Note that all features are all processed in the same way, \emph{individually}, and the ACN block is the \emph{only} place where they interact with each other -- this architecture guarantees permutation-equivariance.
}
\vspace{-.9em}
\label{fig:network}
\end{figure*}

\paragraph{Generating attention.}
We leverage two different types of attention mechanisms, \textit{local} and \textit{global}:
\begin{align}
\w^\text{local}_i = \attnet^\text{local}(\f_i) &= \text{sigmoid}(\bW\f_i^\top + \bb),
\label{eq:local}
\\
\w^\text{global}_i = \attnet^\text{global}(\f_i) &= \frac{\exp\left(\bW\f_i^\top + \bb\right)}{\Sigma_{j=1}^N\exp\left(\bW \f_j^\top + \bb \right)},
\label{eq:global}
\end{align}
where $\bW$ and $\bb$ are the parameters of a perceptron, and $\f_k$ denotes the feature vector for data point~$k$~--~the $k$-th row of the feature map $\f$.
Observe that the \textit{local} attention mechanism~\eq{local} acts on each feature vector \textit{independently}, whereas the \textit{global} attention mechanism~\eq{global} relates the feature vector for each data point to the \textit{collection} through softmax.

\paragraph{Blending attention.}
Note that the product does not change the scale of the normalization applied in \eq{acn}.
Therefore, to take into account multiple types of attention simultaneously, we simply merge them together through element-wise multiplication.
One could use a parametric form of attention blending instead; however, it is non-trivial to combine the weights in a permutation-equivariant way, and we found this simple strategy effective.

\paragraph{Supervising attention.}
In some problems, the class for each data point is known {\em a priori} and explicit supervision can be performed.
In this case, adding a supervised loss on the attention signals can be beneficial.
For instance, when finding good correspondences for stereo we can apply binary-cross entropy using the epipolar distance to generate labels for each putative correspondence, as in~\cite{Yi18a}.
Our experiments in \Section{ablation} show that while this type of supervision can provide a small boost in performance (1-2\%), our approach performs nearly as well without this supervision.

\section{Network architecture and applications}
\label{sec:net}
Our network receives as input $\mathbf{P} {\in} \mathbb{R}^{N \times D}$, the tensor representation of $\mathcal{P}$, and produces an output tensor $\bO {\in} \mathbb{R}^{N \times C}$.
Note that as $\mathcal{P}$ is unstructured, $\bO$ must be equivariant with respect to permutations of the $N$ rows of $\mathbf{P}$.
This output tensor is then used in different ways according to the task at hand.
We model our architecture after~\cite{Yi18a}, which we refer to as {\em Context Network} (CNe).
It features a series of residual blocks~\cite{He16} with Context Normalization (CN). 
Our architecture, which we call {\em Attentive Context Network}, or ACNe, is pictured in~\Figure{network}.
A key distinction is that within each normalization block (\Figure{network}; right) we link the individual outputs of each perceptron $\mathcal{F}_\varphi$ to our ACN layer.
We also replace the Batch Normalization layers~\cite{Ioffe15} used in~\cite{Yi18a} with Group Normalization~\cite{Wu18}, as we found it performs better;~see~\secref{ablation} for ablation tests.

We demonstrate that ACNe can be used to solve multiple applications, ranging from classical problems such as robust line fitting~(\Section{fitting}) and point cloud classification on MNIST and ModelNet40~(\Section{classification}),
to robust camera pose estimation for wide-baseline stereo~(\Section{stereo}).
\begin{figure*}
\centering
\includegraphics[width=0.95\linewidth, trim = 0 0 0 15, clip]{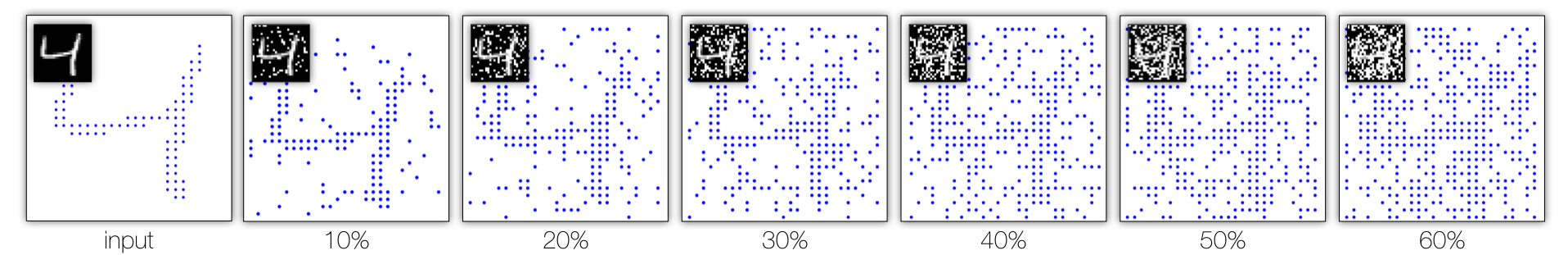}
\caption{
\textbf{Classification} --
We add salt-and-pepper noise to MNIST images, and then 
convert the digits to an unstructured point cloud.
The \% reports the outlier-to-inlier ratio.
}
\vspace{-1.0em}
\label{fig:mnist}
\end{figure*}
\subsection{Robust line fitting}
\label{sec:fitting}
We consider the problem of fitting a line to a collection of points $\mathbf{P}{\in}\mathbb{R}^{N \times 2}$ that is ridden by noise and outliers; see \Figure{toy}.
This problem can be addressed via smooth~(IRLS) or combinatorial~(RANSAC) optimization -- both methods can be interpreted in terms of sparse optimization, such that inliers and outliers are clustered separately; see~\cite{sparseicp}.
Let us parameterize a line as the locus of point $[x,y]$ such that $\boldsymbol{\theta} \cdot [x,y,1]{=}0$.
We can then score each row of $\bP$ (i.e. each 2D point) by passing the output tensor $\bO{=}\text{ACNe}(\bP)$ to an additional weight network~--~with local and global components~--~following \eq{w}, yielding weights $\bw=\normalize(\mathcal{W}_\bomega(\bO))$.
Given $\bw$, and expressing our points in homogeneous coordinates as $\bar\bP=[\bP, 1]{\in}\mathbb{R}^{N \times 3}$, we can compute our covariance matrix as $\mathbf{C}_\bw(\bP){=}\bar\bP^\top \text{diag}(\bw)^2 \bar\bP{\in}\mathbb{R}^{3\!\times\!3}$.
Then, denoting $\nu_0[\mathbf{C}]$ as the eigenvector of $\mathbf{C}$ corresponding to its smallest eigenvalue, $\nu_0\left[\mathbf{C}_{\bw}(\bP)\right]$ is the estimated plane equation that we seek to find.
We, therefore, minimize the difference between this eigenvector and the ground truth, with additional guidance to $\bw^{\text{local}}$ to help convergence:
\begin{align}
\mathcal{L}(\bomega) 
    &= 
    \alpha \min_{+/-}\left\{
    \left\| \nu_0\left[\mathbf{C}_{\bw}(\bP)\right] \pm \boldsymbol{\theta}\ \right\|_2^2
    \right \}
    \nonumber\\
    &\qquad
    + \beta\IE\left[H(\by, \bw^{\text{local}})\right]
,
\label{eq:fitting}
\end{align}
where $\IE\left[H(\ba,\bb)\right]$ is the average binary cross entropy between $\ba$ and $\bb$, $\by$ is the ground-truth inlier label, and hyper-parameters $\alpha$ and $\beta$ control the influence of these losses.
The $\min_{+/-}$ resolves the issue that $-\btheta$ and $\btheta$ are the same line.

\subsection{Point cloud classification}
\label{sec:classification}
We can also apply ACNe to point cloud {\em classification} rather than reasoning about individual points.
As in the previous application, we consider a set of 2D or 3D locations $\mathbf{P} {\in} \mathbb{R}^{N \times D}$ as input, where $D$ is the number of dimensions.
In order to classify each point set, we transform the output tensor $\bO{=}\text{ACNe}(\bP)$ into a single vector $\bv{=}\mu_\bw\left(\bO\right)$ and associate it with a ground-truth one-hot vector $\by$ through $\softmax$.
Additional weight networks to generate $\bw$ are trained for this task.
We train with the cross entropy loss.
Thus, the loss that we optimize is:
\begin{equation}
\mathcal{L}(\bomega) = H\left(\by, \softmax(\bv)\right)
.
\label{eq:classification}
\end{equation}

\subsection{Wide-baseline stereo}
\label{sec:stereo}
In stereo we are given correspondences as input, which is thus
$\bP\in\IR^{N\times4}$, where $N$ is the number of correspondences and each row contains two pixel locations on different images.
In order to remain comparable with traditional methods, we aim to solve for the {\em Fundamental} matrix, instead of the {\em Essential} matrix, \ie, without assuming known camera intrinsics.
Thus, differently from~\cite{Yi18a,Zheng18,Zhao19}, we simply normalize the image coordinates with the image size instead.
This makes our method more broadly applicable, and directly comparable with most robust estimation methods for stereo~\cite{Fischler81,Torr00,Chum03,Chum05b,Barath19}.

We obtain $\bw$ from the output tensor $\bO {=} \text{ACNe}(\bP)$ via \eq{w}, as in \Section{fitting}.
The weights $\bw$ indicate which correspondences are considered to be inliers and their relative importance.
We then apply a \textit{weighted} variant of the 8-point algorithm~\cite{Hartley00} to retrieve the Fundamental matrix~$\hat{\bF}$, which parameterizes the relative camera motion between the two cameras.
To do so we adopt the differentiable, non-parametric form proposed by \cite{Yi18a}, and denote this operation as~$\hat{\bF} {=} g\left(\bX, \bw \right)$.
We then train our network to regress the ground-truth Fundamental matrix, as well as providing auxiliary guidance to  $\bw^{\text{local}}$~--~the \textit{final} local attention used to construct the output of the network~--~with per-correspondence labels obtained by thresholding over the symmetric epipolar distance~\cite{Hartley00}, as in \cite{Yi18a}.
In addition, we also perform auxiliary supervision on $\bw_k^{\text{local}}$~--~the \textit{intermediate} local attentions within the network~--~as discussed in \Section{acn}. Note that this loss is not necessary, but helps training and provides a small boost in performance; see~\Section{ablation}.
We do not supervise global attention and leave it for the network to learn.
We therefore write:
\begin{align}
    \loss(\bomega) 
    &=
    \alpha\min_{+/-}\left\{
        \left\|\hat{\bF} \pm \bF^*\right\|_F^2
        \right\}
    + \beta\IE\left[H(\by, \bw^{\text{local}})\right] 
    \nonumber \\
    &\qquad
    + \gamma\IE_k\left[H(\by, \bw_k^{\text{local}})\right]
    ,
    \label{eq:stereo}
\end{align}
where $\|\cdot\|_F$ is the Frobenius norm, $H$ is the binary cross entropy, and $\by$ denotes ground truth inlier labels. 
Again, the hyper-parameters $\alpha$, $\beta$, and $\gamma$ control the influence of each loss.
Similarly to the line-fitting case, the $\min_{+/-}$ resolves the issue that $-\bF^*$ and $\bF^*$ express the same solution.

\section{Implementation details}
\label{sec:impl}
We employ a K-layer structure (excluding the first linear layer that changes according to the number of channels) for ACNe, with K$\times$ ARB units, and two perceptron layers in each ARB.
The number of layers K is set to 3$\times$ for
2D point cloud classification, 
6$\times$ for robust line fitting, and 12$\times$ for stereo.
For 3D point cloud classification, we add ACN normalization to an existing architecture.
We also use 32 groups for Group Normalization, as suggested in~\cite{Wu18}. Similarly to~\cite{Yi18a}, we use $C{=}128$ channels per perceptron.

\paragraph{Training setup.}
For all applications we use the ADAM optimizer~\cite{Kingma15} with default parameters and a learning rate of $10^{-3}$.
Except for robust line fitting, we 
use a validation set to perform early stopping.
For robust line fitting, the data is purely synthetic and thus infinite, and we
train for 50k iterations.
For MNIST, we use 70k samples with a 8:1:1 split for training, validation and testing.
For stereo, we use the splits from~\cite{Zhang19a}.
For the loss term involving eigen-decomposition (terms multiplied by $\alpha$ in~\eq{fitting} and~\eq{stereo}), we use $\alpha{=}0.1$, following~\cite{Yi18a}. 
All other loss terms have a weight of $1$, that is, $\beta{=}1$ and $\gamma{=}1$.
For stereo, we follow \cite{Yi18a} and enable the term involving the Fundamental matrix -- the first term in \eq{stereo} -- after 20k iterations.

\paragraph{Robust estimators for stereo inference.}
As a special case, we evaluate the possibility
of applying standard robust estimators for outlier rejection such as RANSAC after training the model to potentially maximize its performance, as previously done in~\cite{Yi18a,Zheng18,Zhang19a}.
To do so, we modify our architecture by changing the final layer to output only the local attention with the ReLU$+$Tanh activation, as in Yi~et~al.~\cite{Yi18a}.
We then simply threshold~$\bw$ with zero, select the data points that survive this process as inliers, and feed them to different RANSAC methods to process them further.
We compare these results with those obtained directly from the weighted 8-point formulation.

\section{Results}
\label{sec:result}

We first consider a toy example on fitting 2D lines with a large ratio of outliers.
We then apply our method to point cloud classification, following~\cite{Qi17a,Qi17b}, which includes 2D for digit classification on MNIST and 3D for object classification on ModelNet40~\cite{Wu15b}.
These three experiments illustrate that our attentional method performs better than vanilla Context Normalization under the presence of outliers. 
We then apply our solution to wide-baseline stereo, and demonstrate that this increase in performance holds on challenging real-world applications, and against state-of-the-art methods for robust pose estimation. 
Finally, we perform an ablation study and evaluate the effect of supervising the weights used for attention in stereo.

\begin{table}
\scriptsize
\begin{center}
\setlength{\tabcolsep}{11pt}
\begin{tabular}{lccccc}
\toprule
Outlier ratio & 60\% & 70\% & 80\% & 85\% & 90\% \\
\midrule

CNe~\cite{Yi18a} & .00019& .0038 & .056 & .162 & .425\\
ACNe (Ours) & {\bf1e-6} & {\bf .0008} & {\bf .024} & {\bf .130} & {\bf .383} \\
\bottomrule
\end{tabular}
\end{center}
\vspace{-.5em}
\caption{
{\bf Robust line fitting} -- Line fitting results over the test set in terms of the $\ell_2$ distance
between ground-truth and the estimates.
}
\label{tbl:linefitting}
\end{table}

\begin{table}
\scriptsize
\begin{center}
\setlength{\tabcolsep}{7pt}
\begin{tabular}{lccccccc}
\toprule
Outlier ratio & 0\% & 10\% & 20\% & 30\% & 40\% & 50\% & 60\% \\
\midrule

PointNet~\cite{Qi17a} & 98.1 & 95.1 & 93.2 & 79.5 & 67.7 & 70.0 & 54.8 \\
CNe~\cite{Yi18a} & 98.0 & 95.8 & 94.0 & 91.0 & 90.1 & 87.7 & 87.2 \\
ACNe (Ours) & \textbf{98.3} & \textbf{97.2} & \textbf{96.5} & \textbf{95.3} & \textbf{94.7} & \textbf{94.3} & \textbf{93.7} \\

\bottomrule
\end{tabular}
\end{center}
\vspace{-.5em}
\caption{
{\bf 2D Point cloud classification} -- Classification accuracy on MNIST, under different outlier ratios (\%).
Our method performs best in all cases, and the gap becomes wider with more outliers.
}
\vspace{-.7em}
\label{tbl:mnist}
\end{table}

\subsection{Robust line fitting -- \Fig{toy} and \Table{linefitting}}
\label{sec:toy}
To generate 2D points on a random line, as well as outliers, we first sample 2D points uniformly within the range $[-1, +1]$.
We then select two points randomly and fit a line that goes through them.
With probability according to the desired inlier ratio, we then project each point onto the line to form inliers.
We measure the error in terms of the $\ell_2$ distance between the estimated and ground truth values for the line parameters.
The results are summarized in \tbl{linefitting}, with qualitative examples in \Figure{toy}.
ACNe consistently outperforms CNe~\cite{Yi18a}.
Both methods break down at a 85-90\% outlier ratio, while the performance of ACNe degrades more gracefully.
As illustrated in \Figure{toy}, our method learns to progressively focus on the inliers throughout the different layers of the network and weeds out the outliers.

\subsection{Classifying digits -- \Fig{mnist} and \Table{mnist}}
We evaluate our approach on handwritten digit classification on MNIST, which consists of $28 \times 28$ grayscale images. 
We create a point cloud from these images following the procedure of \cite{Qi17a}: we threshold each image at 128 and use the coordinates~--~normalized to a unit bounding box~--~of the surviving pixel locations as data samples.
We subsample 512 points with replacement, in order to have the same number of points for all training examples.
We also add a small Gaussian noise of 0.01 to the pixel coordinates after sampling following~\cite{Qi17a}.
Outliers are generated by sampling from a uniform random distribution. 
We compare our method against vanilla PointNet~\cite{Qi17a} and CNe~\cite{Yi18a}.
For PointNet, we re-implemented their method under our framework to have an identical training setup.

\Table{mnist} summarizes the results in terms of classification accuracy.
Our method performs best, with the gap widening as the outlier ratio increases -- while CNe shows some robustness to noise, PointNet quickly breaks down.
Note that the results for PointNet are slightly different from the ones reported in \cite{Qi17a}, as we use a validation split to perform early stopping.
In addition, to reduce randomness, we train 10 different models and report the average results.

\begin{table}
    \scriptsize
    \begin{center}
    \setlength{\tabcolsep}{8pt}
    \begin{tabular}{lcccccc}
    \toprule
    Outlier ratio & 0\% & 10\% & 20\% & 30\% & 40\% & 50\% \\ %
    \midrule
    PointNet          & 85.8	& 81.7	&81.7	&80.1	&78.2	&78.5	\\% &76.7 \\
    PointNet w/ CN    & 87.2	& 84.3	&84.5	&83.4	&81.7	&81.6	\\% &81.5 \\
    PointNet w/ ACN   & {\bf 87.7}	& {\bf 84.6}	& {\bf 85.0}	& {\bf 84.6}	&{\bf 83.3}	& {\bf 84.2}	\\% &84.1\\
    \bottomrule
    \end{tabular}
    \end{center}
    \vspace{-0.5em}
    \caption{{\bf 3D Point cloud classification --} We replicate the 3D point classification experiment on ModelNet40 from~\cite{Qi17a}, with {\em vanilla} PointNet.
    We then add outliers with Gaussian noise.
    Our approach performs best with and without outliers.}
    \label{tbl:pointcloud3d}    
    \vspace{-0.7em}
\end{table}

\newcommand{\imheight}{.20\linewidth}
\newcommand{\imindexa}{1}
\newcommand{\imindexb}{6}
\begin{figure*}
\centering
\setlength{\tabcolsep}{0.25pt}
\begin{tabular}{@{}cccccc@{}}
\includegraphics[height=\imheight]{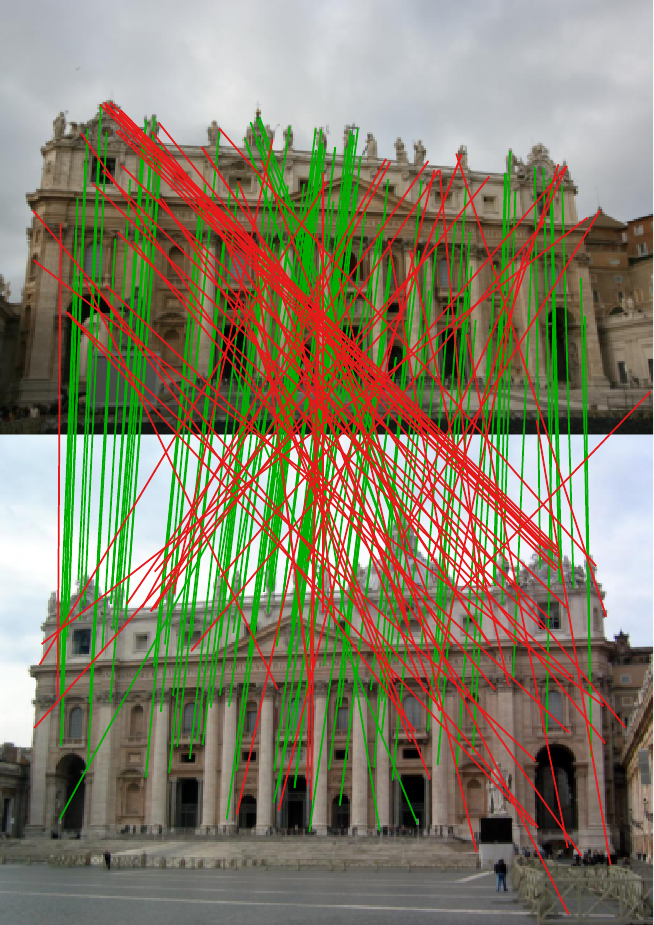} &
\includegraphics[height=\imheight]{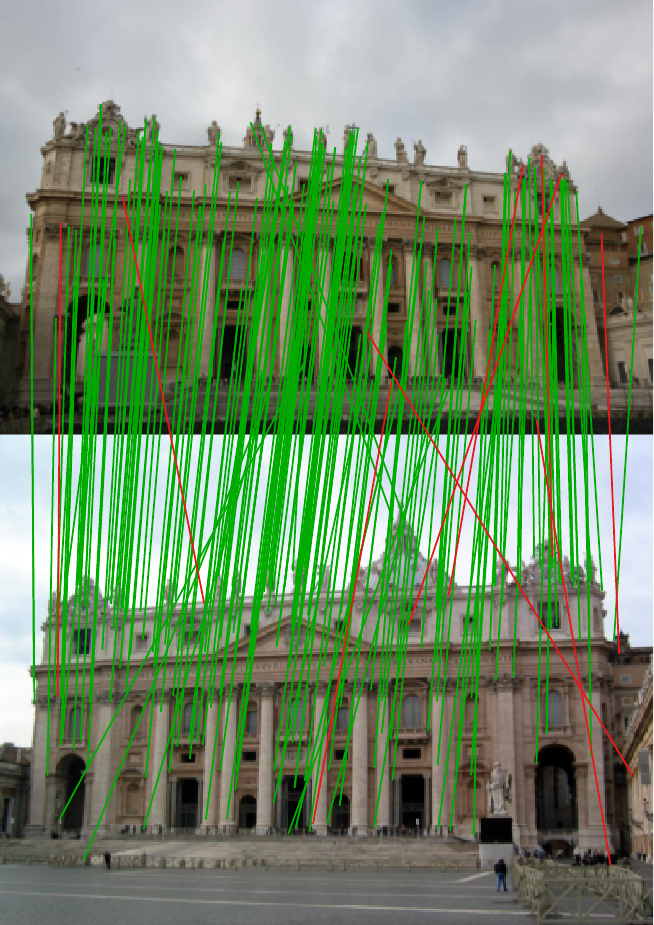} &
\includegraphics[height=\imheight]{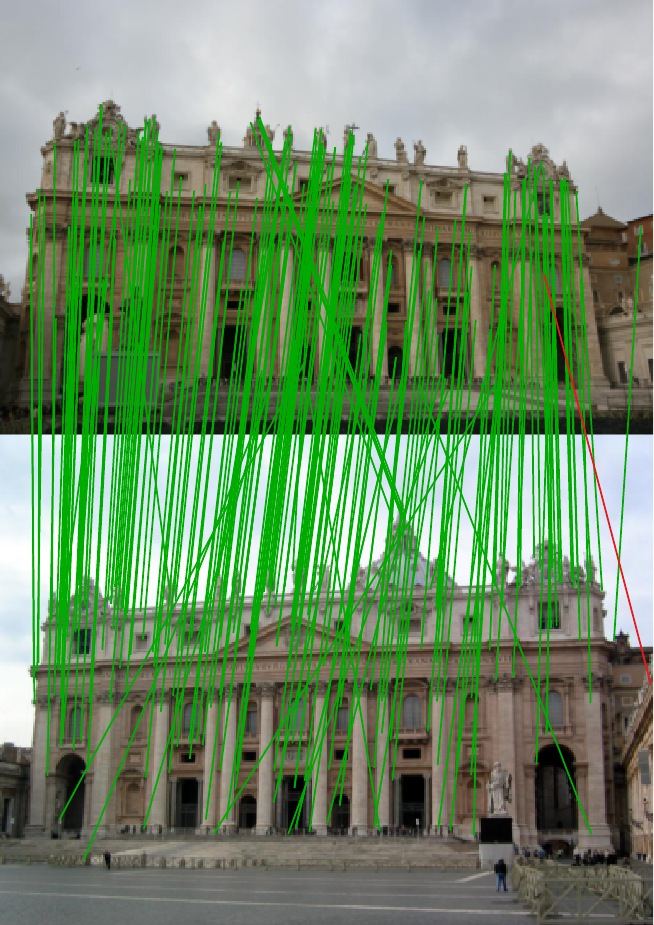} &
\includegraphics[height=\imheight]{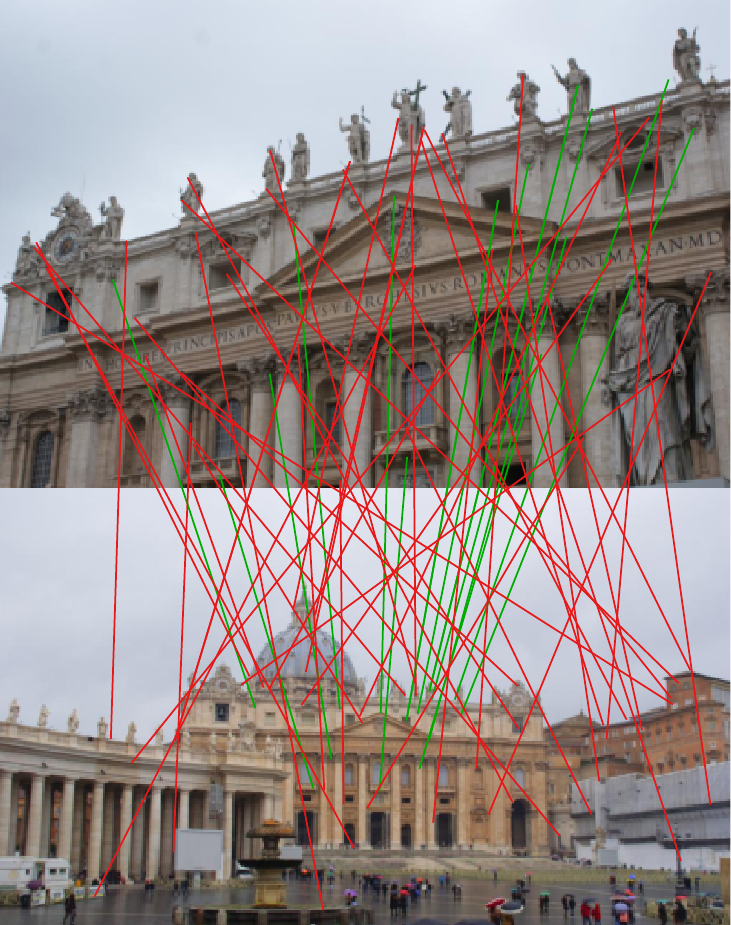} &
\includegraphics[height=\imheight]{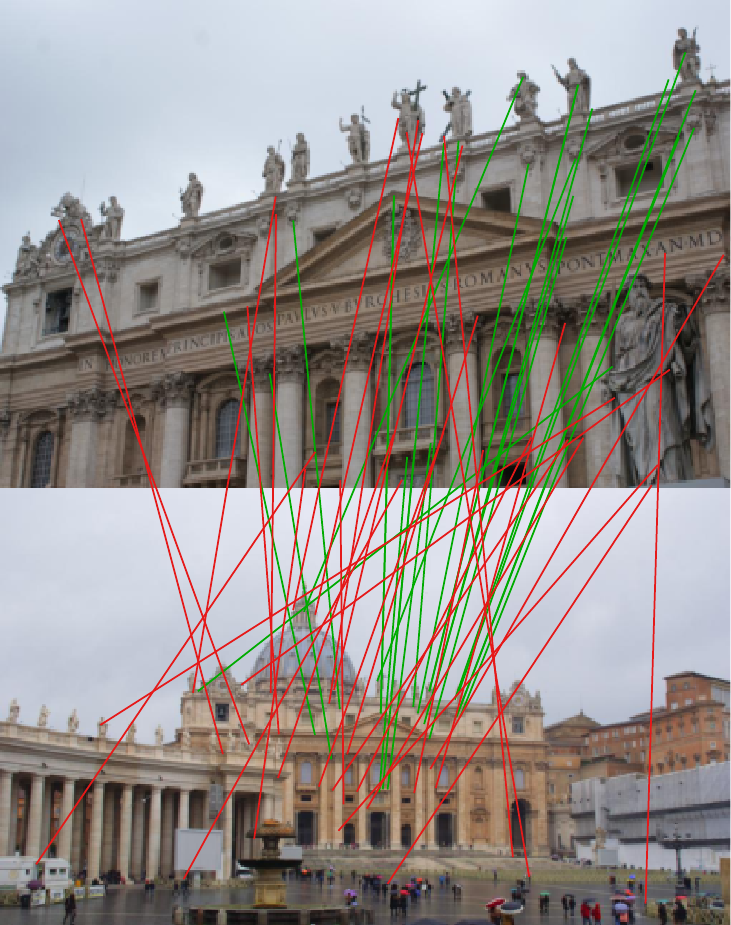} &
\includegraphics[height=\imheight]{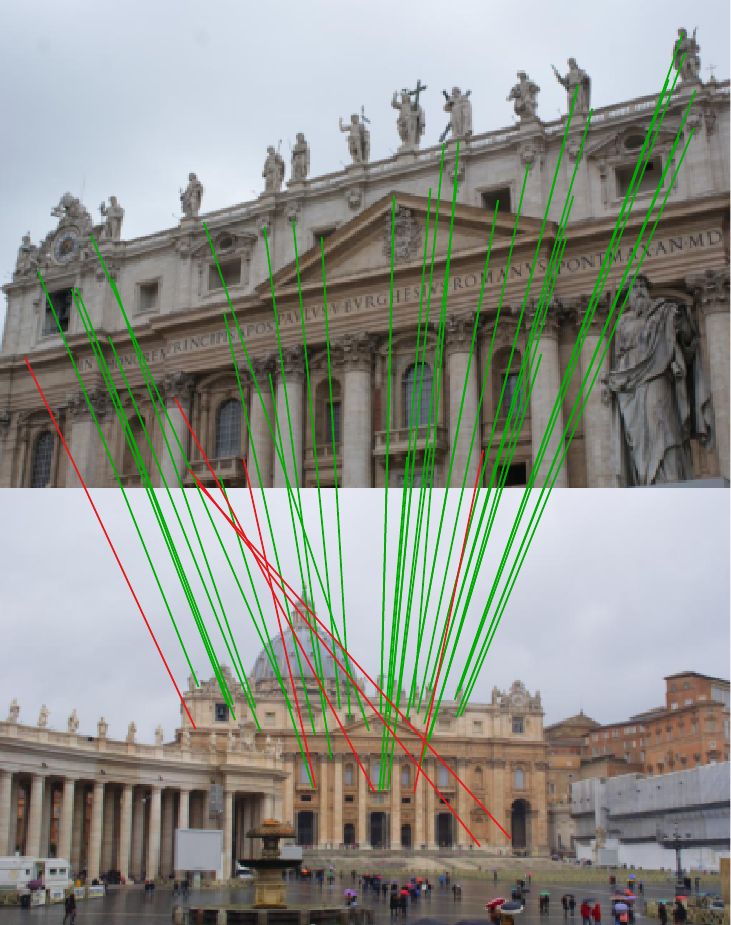} \\
{\scriptsize (a) RANSAC~\cite{Fischler81}} & {\scriptsize (b) CNe~\cite{Yi18a}} & {\scriptsize (c) ACNe (ours)} & {\scriptsize (d) RANSAC~\cite{Fischler81}} & {\scriptsize (e) CNe~\cite{Yi18a}} & {\scriptsize (f) ACNe (ours)} \\
\end{tabular}
\vspace{2mm}
\caption{
{\bf Wide-baseline stereo} --
We show the results of different matching algorithms on the dataset of~\cite{Yi18a}. We draw the {\em inliers} produced by them, in {\color{dark_green}{\bf green}} if the match is below the epipolar distance threshold (in {\color{red}{\bf red}} otherwise).
Note that this may include some false positives, as epipolar constraints map points to lines -- perfect ground truth would require dense pixel-to-pixel correspondences.
}
\label{fig:stereo}
\end{figure*}

\subsection{Classifying 3D objects -- \Table{pointcloud3d}}
We apply our method to the problem of 3D object (point cloud) classification.
We use the ModelNet40 dataset~\cite{Wu15b}, and compare with PointNet~\cite{Qi17b}.
Similarly to the MNIST case, we contaminate the dataset with outliers to test the robustness of each method.
Specifically, we add a pre-determined ratio of outliers to the point clouds, sampled uniformly within the range $[-1, 1]$. 
We also add small Gaussian perturbations to the locations of the points, with a standard deviation of $0.01$.
We then sample 1024 points from the point cloud to perform classification.
Again, to simply test if ACN can improve existing pipelines, we plug our normalization into the vanilla PointNet architecture.
Note that the original PointNet includes an affine estimation step which provides a small performance boost -- we omit it from our implementation, in order to isolate the architectural differences between the methods.
We report the results in \Table{pointcloud3d}. Our method performs best, with the gap becoming wider as outliers become prevalent.

\subsection{Wide-baseline stereo -- \Fig{stereo} and \Table{stereo}}
\label{sec:quantitative}
Wide-baseline stereo is an extremely challenging problem, due to the large number of variables to account for -- viewpoint, scale, illumination, occlusions, and properties of the imaging device -- see~\Fig{stereo} for some examples.
We benchmark our approach on a real-world dataset against multiple state-of-the-art baselines, following the data~\cite{Zhang19a} and protocols provided by~\cite{Yi18a}.
Their ground truth camera poses are obtained from Structure-from-Motion with VisualSfM~\cite{Wu13}, from large image collections of publicly available, challenging photo-tourism data.

We evaluate performance in terms of the {\em reconstructed poses}.
Since the stereo matching problem is defined only up to a scale factor~\cite{Hartley00}, it is not possible to compute absolute (metric) errors for translation.
Instead, we follow the methodology of~\cite{Yi18a} and measure the error between the ground truth and estimated vectors between both cameras, for both rotation and translation, and combine them by taking the maximum of the two.
We then evaluate the accuracy over all image pairs at multiple error thresholds, accumulate it up to a limit (either 10$^\circ$ or 20$^\circ$), and summarize performance by its mean -- which we call the mean Average Precision (mAP); see \cite{Yi18a}.
This means that methods that perform better at lower error thresholds are rewarded.
We use their data mostly as is, using the pre-extracted correspondences and splits from OANet~\cite{Zhang19a}, but adapt it to the Fundamental matrix problem.
In contrast to previous works~\cite{Yi18a,Zhang19a}, which report results on the scene the models were trained on, we focus on unknown scenes, in order to determine each method's {\em actual} performance.

As we discussed in \secref{stereo}, both CNe~\cite{Yi18a} and OANet~\cite{Zhang19a} assume known camera intrinsics and estimate the Essential matrix, instead of the Fundamental matrix -- this is a \textit{significantly} easier problem, as the number of free parameters drops from 7 to 5.
However, most research papers on this topic focus on estimating the Fundamental matrix~\cite{Chum03,Chum05b,Raguram12,Barath18,Barath19},
which is why we focus on this problem instead.
For completeness, we also report results for the Essential matrix in the supplementary appendix, for which we also achieve state-of-the-art results.

In more detail, given an image pair, we extract 2k keypoints for each image with SIFT~\cite{Lowe04}. 
Matches are then formed from one image to the other, in both directions.
As is typical for image matching, we then filter out non-discriminative correspondences via bi-directional check, enforcing one-to-one matching.
For RANSAC variants we found it to be critical to further apply Lowe's ratio test~\cite{Lowe04}~--~without it RANSAC variants provide worse results.
We apply it with a ratio threshold of 0.8.
We do not apply this test for learned methods, as it throws out too many inliers for learned methods to bring any benefit.
Also, when training learned methods, we train without bidirectional check to show as many correspondences in the training set as possible to the network.

We consider the following methods: LMedS~\cite{Rousseeuw84}, RANSAC~\cite{Fischler81,Chum03}, MLESAC~\cite{Torr00}, DegenSAC~\cite{Chum05b}, MAGSAC~\cite{Barath19} CNe~\cite{Yi18a}, DFE~\cite{Ranftl18}, OANet~\cite{Zhang19a} and ACNe (ours).
We consider the pose estimated with the weighted 8-point algorithm directly, as well as those combined with a robust outlier rejection method as outlined in~\refsec{impl}.

\paragraph{Quantitative results.}
We report quantitative results in \Table{stereo}, for two different error thresholds (10$^\circ$/20$^\circ$).
We make three fundamental observations:

\noindent
\CIRCLE{1} Our method consistently outperforms all of the baselines, including CNe and OANet. 
The difference in performance between ACNe and its closest competitor, OANet, is of $14.1/9.8\%$ relative for Outdoors, and $8.6/9.2\%$ relative for Indoors when used without any additional post processing.
The gap for Outdoors is reduced to $1\%$ 
when they are combined with MAGSAC, but ACNe still outperforms OANet.
For Indoors, we observe a drop in performance for both OANet and ACNe when combining them with RANSAC or MAGSAC.
The margin between learned and traditional methods is significant, with ACNe performing $30.1/39.6\%$ better relative on Outdoors and $45.5/47.1\%$ better relative on Indoors, compared to the best performing traditional baseline -- including a very recent method, MAGSAC.

\begin{table}
\begin{center}
\scriptsize
\setlength\tabcolsep{13pt} %
\begin{tabular}{clcc}
\toprule
& Method & {Outdoors} & {Indoors}\\
\midrule
\multirow{5}{*}{\rotatebox[origin=c]{90}{Traditional}} & LMedS & {.296/.383} &  {.142/.235}  \\
& RANSAC &  {.356/.437}  & {.172/.272}  \\
& MLESAC &  {.148/.216}  & {.135/.230}  \\
& DegenSAC &  {.328/.394}  & {.191/.291}  \\
& MAGSAC &  {.385/.457}  & {.185/.282}  \\
\midrule
\multirow{9}{*}{\rotatebox[origin=c]{90}{Learned}} 
& CNe (weighted-8pt) &  {.323/.469}  & {.189/.331}  \\
& CNe+RANSAC &  {.449/.554}  & {.201/.315}  \\
& CNe+MAGSAC &  {.500/.598}  & {.213/.326}  \\
& DFE (weighted-8pt)\footnotemark &  {.319/.470}  & {.167/.294}  \\
& DFE+RANSAC &  {.414/.508}  & {.193/.303}  \\
& DFE+MAGSAC &  {.452/.541}  & {.211/.320}  \\
& OANet (weighted-8pt) &  {.439/.581}  & \underline{.256/.392}  \\
& OANet+RANSAC &  {.482/.592}  & {.211/.331}  \\
& OANet+MAGSAC &  \underline{.514}{/.615}  & {.230/.346}\\
\midrule
\multirow{3}{*}{\rotatebox[origin=c]{90}{Ours}} 
& ACNe (weighted-8pt) & {.501/}{\bf .638} & {\bf .278/.428} \\
& ACNe+RANSAC & {.478/.590} & {.209/.329} \\
& ACNe+MAGSAC & {\bf .518}\underline{/.621}& {.226/.343} \\

\bottomrule
\end{tabular}
\end{center}
\vspace{-0.3em}
\caption{
{\bf Pose estimation accuracy} -- mAP at 10$^\circ$/20$^\circ$ error threshold.
Similarly to \cite{Yi18a}, we consider multiple baselines, as well as pairing different methods with state-of-the-art RANSAC variants.
Our method consistently outperforms all others
by a significant margin, even without an additional robust estimator, in some cases.
} %
\vspace{-0.7em}
\label{tbl:stereo}
\end{table}

\noindent
\CIRCLE{2} Different from the findings of~\cite{Yi18a}, we observe that RANSAC variants may harm performance, particularly with ACNe.
This is because through its global attention -- $\bw^{\text{global}}$ -- ACNe can infer the relative importance of each correspondence, which is not easily taken into account when passing samples to a robust estimator.
In this manner, ACNe goes {\em beyond} simple outlier rejection.
The best performance is typically achieved by using ACNe at its pure form, directly feeding its weights to the weighted 8-point algorithm.
Given that all our experiments are on unseen sequences, this further shows that ACNe generalizes very well,
even without being followed by an additional robust estimator.

\noindent
\CIRCLE{3} Contrary to the results of Yi~\etal~\cite{Yi18a} and Zhang~\etal~\cite{Zhang19a}, we find that traditional baselines perform better than reported on either work.
This is because their experimental setup did not consider Lowe's ratio test, nor the bidirectional check.
Without these, the performance of traditional baselines drops drastically
-- RANSAC and MAGSAC drop $79.2/73.0\%$ and $92.0/85.1\%$ relative performance, respectively, for Outdoors, and $66.9/59.2\%$ and $82.2/74.1\%$ for Indoors.

\footnotetext{DFE, when tested with bidirectional check, gives poor results -- .236 for Outdoors (20$^o$). We hypothesize this may be due to the fact that DFE uses additional, ``side'' information, whose statistics may vary drastically with the bidirectional check.
We therefore do not use it for this baseline. This problem disappears when additional robust estimators are used.}

\paragraph{Ablation study -- \Table{ablation}.}
\label{sec:ablation}
We perform an ablation study to evaluate the effect of the different types of attention, as well as the supervision on the local component of the attentive mechanism. 
We also compare with CNe, as its architecture is the most similar to ours.
We use the train and validation splits for the \emph{Saint Peter's Square} sequence for this study, as it is the primary sequence used for training in \cite{Yi18a} and has many images within the set.
\CIRCLE{1} We confirm that CNe~\cite{Yi18a} performs better with Batch Normalization (BN)~\cite{Ioffe15} than with Group Normalization (GN)~\cite{Wu18} -- we use GN for ACNe, as it seems to perform marginally better
with our attention mechanism.
\CIRCLE{2} We observe that our attentive mechanisms allow ACNe to outperform CNe, and that their combination outperforms their separate use.
\CIRCLE{3}~Applying supervision on the weights further boosts performance.

\paragraph{With learned features -- \Table{feats}.} 
Finally, we report that our method also works well with two state-of-the-art, learned local feature methods -- SuperPoint~\cite{DeTone17b} and LF-Net~\cite{Ono18}.
They are learned end-to-end -- their characteristics are thus different from those of SIFT keypoints.
We test again on \textit{Saint Peter's Square}, as our primary focus is to show that it is possible to use other feature types.
In \Table{feats} we report that both methods improve performance over SIFT with OANet and ACNe, but with ratio test and MAGSAC they perform worse.
It is interesting how SuperPoint, without the ratio test, performs better than SIFT with MAGSAC, but the order is reversed when the ratio test is introduced, highlighting its importance.
Regardless of feature type, we demonstrate that our approach provides improved performance over other methods, and that it pairs best with SuperPoint.

\begin{table}
\begin{center}
\footnotesize
\setlength\tabcolsep{6pt} %
\begin{tabular}{lcc cccc}
\toprule
\multirow{2}{*}{Methods} & \multicolumn{2}{c}{CNe~\cite{Yi18a}} & \multicolumn{4}{c}{ACNe (Ours)} \\
\cmidrule(r){2-3}
\cmidrule(r){4-7}
& w/ BN & w/ GN & L & G & L+G & L+G+S \\
\midrule

Weighted-8pt & {.435} & {.414} & {.531} & {.593} & {.597} & {\bf .602}  \\
\bottomrule
\end{tabular}
\end{center}
\caption{
{\bf Ablation study} -- mAP at 20$^\circ$ with different CNe~\cite{Yi18a} and ACNe (ours) variants on stereo.
The labels indicate: {\em L} -- local attention; {\em G} -- global attention; {\em S} -- local attention supervision.
}
\label{tbl:ablation}
\vspace{-0.5em}
\end{table}

\begin{table}
\begin{center}
\footnotesize
\setlength\tabcolsep{10pt} %
\begin{tabular}{lccc}
\toprule
 & SIFT & SuperPoint & LF-Net \\
\midrule

MAGSAC (w/o ratio test) & {.146} & {.205} & {.134} \\
MAGSAC & {.264} & {.230} & {.157} \\
OANet (weighted 8pt) & \underline{.488} & \underline{.547} & \underline{.543} \\
OANet+MAGSAC & {.479} & {.442} & {.452} \\
\midrule
ACNe (weighted 8pt) & {\bf .602} & {\bf .637} & {\bf .619} \\

\bottomrule
\end{tabular}
\end{center}
\caption{{\bf With learned local features --} mAP at 20$^\circ$ with learned local features and different methods. Our method outperforms other methods and performs best with SuperPoint~\cite{DeTone17b}. 
} %
\vspace{-.5em}
\label{tbl:feats}
\end{table}

\section{Conclusion}
\label{sec:conclusion}
We have proposed {\em Attentive Context Normalization}~(ACN), and used it to build {\em Attentive Context Networks}~(ACNe) to solve problems on permutation-equivariant data.
Our solution is inspired by IRLS, where one iteratively re-weighs the importance of each sample, via a soft inlier/outlier assignment.
We demonstrated that by learning both local and global attention we are able to outperform state-of-the-art solutions on line fitting, classification of point clouds in 2D (digits) and 3D (objects),
and challenging wide-baseline stereo problems.
Notably, our method \textit{thrives} under large outlier ratios.
For future research directions, we consider incorporating ACN into general normalization techniques for deep learning. 
We believe that this is an interesting direction to pursue, as all existing techniques make use of statistical moments.
\section*{Acknowledgements}
This work was supported by the Natural Sciences and Engineering Research Council of Canada (NSERC) Discovery Grant, NSERC Collaborative Research and Development Grant (Google), and by Compute Canada.

{\small
  \balance
  \bibliographystyle{ieee_fullname}
  \bibliography{abbr,kwang_paper,vision,learning,weiwei,random}
}
\title{
ACNe: Attentive Context Normalization \\ for Robust Permutation-Equivariant Learning
\\[.5em] 
{(Supplementary Material)}}
\date{}
\maketitle

\appendix
\nobalance

\section{Visualizing attention -- \Figure{visattn}}
\label{sec:vis_attn}
We visualize the weights for wide-baseline stereo, along with the ``ground truth'' labels on the matches.
Since the labels are obtained by thresholding the epipolar distance, computed from the ground truth poses, they contain a few false positives. 
This figure shows that ACN learns to focus on inliers by emulating a robust iterative optimization.
As our system is trained by Fundamental matrix supervision, it was also able to learn to ignore these false positives.

\newcommand{\imwidth}{.495\linewidth}
\begin{figure}%
\centering
\setlength{\tabcolsep}{0.25pt}
\begin{tabular}{@{}ccc@{}}
\includegraphics[width=\imwidth]{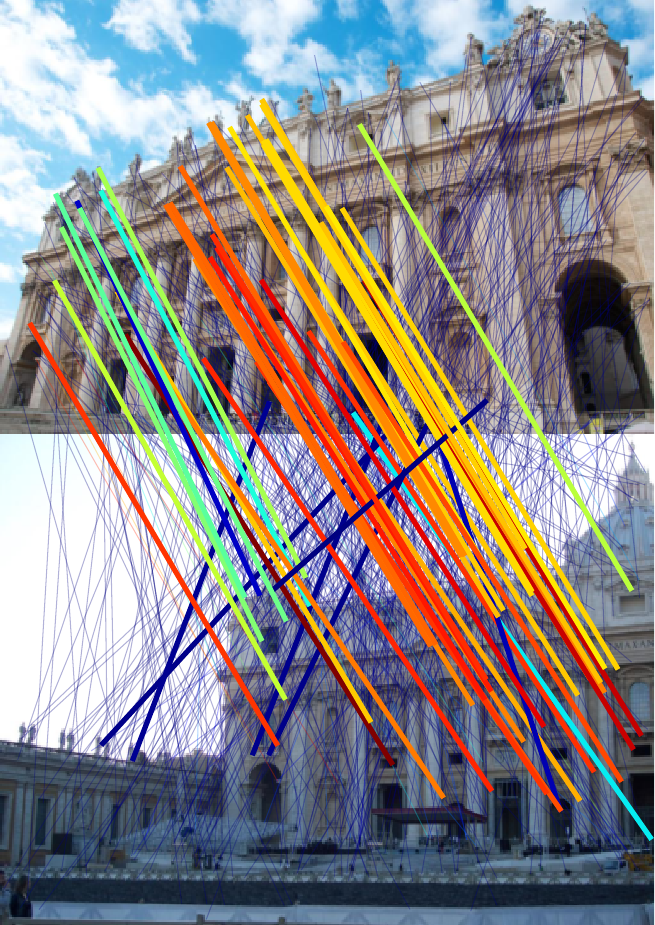} &
\includegraphics[width=\imwidth]{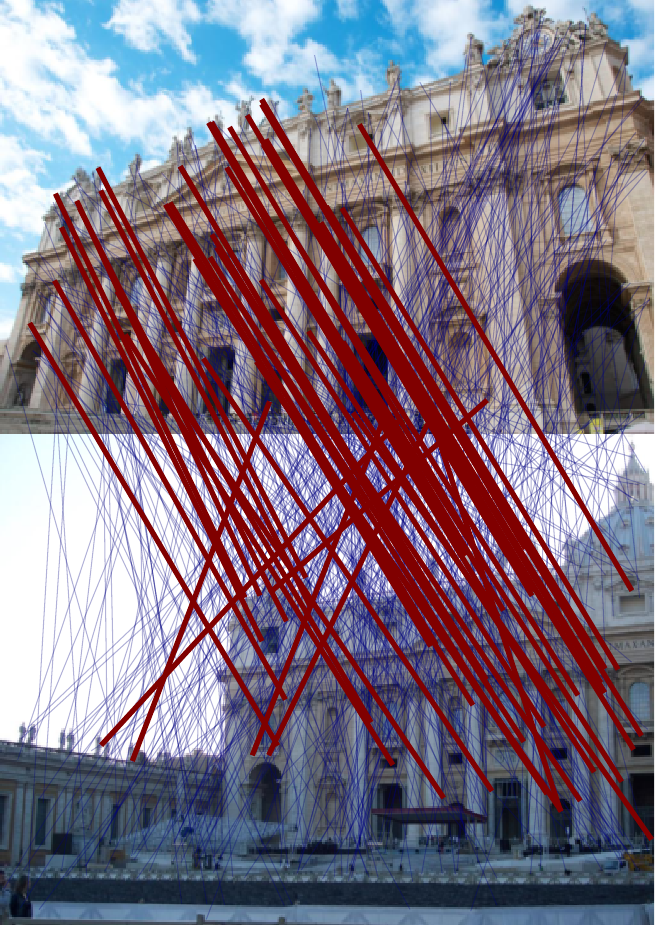} & \\
\multicolumn{2}{c}{{\footnotesize Matches: ACN \hspace{.9in} Matches: Ground Truth }} \\
\multicolumn{2}{c}{\includegraphics[width=0.99\linewidth]{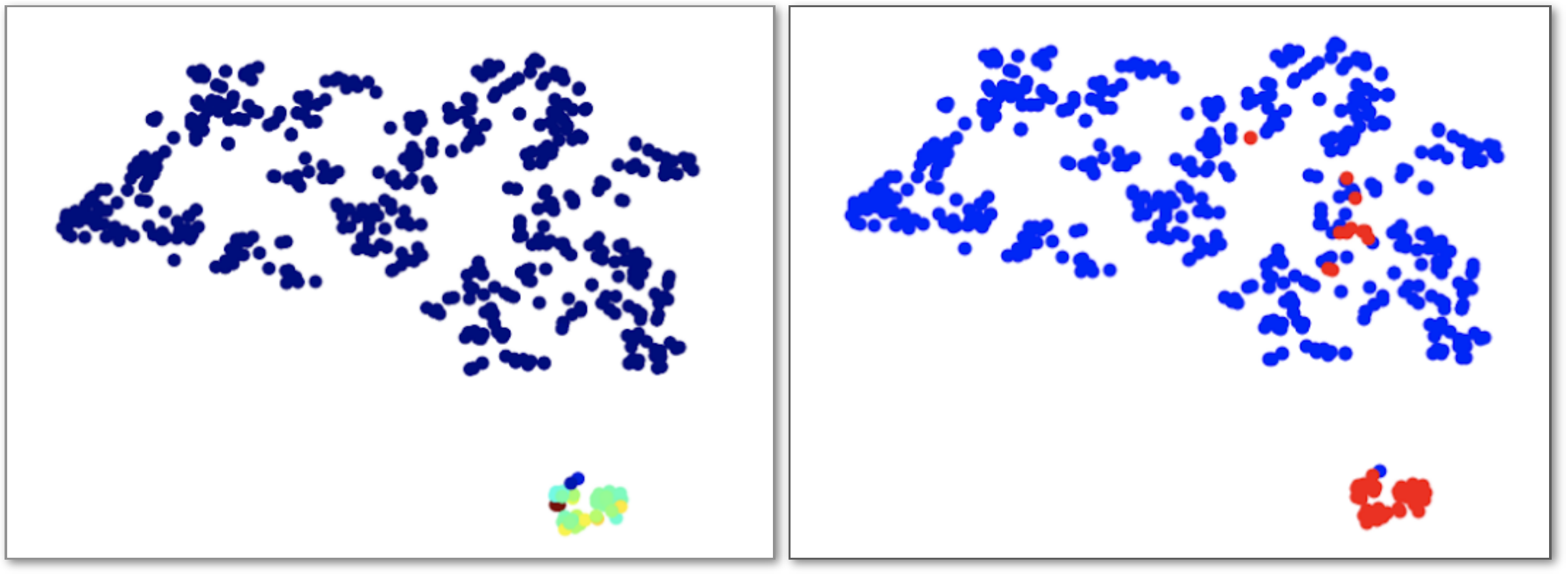}} \\
\multicolumn{2}{c}{{\footnotesize Feature maps (t-SNE): for ACN \hspace{.1in} Feature maps (t-SNE): for GT }} \\
\end{tabular}
\vspace{0mm}
\caption{
{\bf Attention in wide-baseline stereo.}
An illustration of ACNe on an example image pair.
The top row shows the \textit{matches}, and the bottom row shows a representation of the \textit{feature maps} obtained via t-SNE.
The left column displays ACN weights, color-coded by magnitude (highest in {\bf \textcolor{red}{red}}, lowest in {\bf \textcolor{blue}{blue}}), and the right column the ground truth match labels (inliers in {\bf \textcolor{red}{red}}, outliers in {\bf \textcolor{blue}{blue}}), computed by thresholding the symmetric epipolar distance (Sec.~4.3).
We draw matches with negative labels with thinner lines.
}
\label{fig:visattn}
\end{figure}

\begin{table}[tbh!]
\begin{center}
\footnotesize
\setlength\tabcolsep{3pt} %
\begin{tabular}{lccc ccc}
\toprule
\multirow{2}{*}{Methods} & \multicolumn{3}{c}{Attn. on feature map} & \multicolumn{3}{c}{Our method} \\
\cmidrule(r){2-4}
\cmidrule(r){5-7}
& L & G & L+G & L & G & L+G \\
\midrule

Weighted-8pt & .410 & .260 & .408 & {.531}& {.593} & {\bf .597}\\   
Weighted-8pt (ReLU$+$Tanh) & .427 & .347 & .369 &  --- & --- &  --- \\   
\bottomrule
\end{tabular}
\end{center}
\caption{
{\bf Applying attention to the feature maps -- }
We compare our method (right) to applying attention directly to the feature maps (left). 
We report mAP at a 20$^\circ$ error threshold on our validation set -- the \emph{Saint Peter's Square} scene.
Applying attention on the normalization performs significantly better than applying attention to the feature maps.
}
\label{tbl:attfeat}
\end{table}

\section{Attention on feature maps -- \Table{attfeat}}
\label{sec:featmap_vs_norm}

We show that applying attention to the {\em normalization} of the feature maps (our method) outperforms the more commonly used strategy of applying attention directly to the feature maps~\cite{Wang17b,Dusmanu19}.
\Table{attfeat} extends our ablation study from \Table{ablation}, demonstrating that our method outperforms this alternative approach by 29\% relative.

It is important to note that introducing global attention to the feature maps resulted in \textit{unstable} training.
To avoid gradient explosion we reduced the learning rate to one-tenth of the value we typically used.
Nonetheless, gradients exploded after 224k iterations.
We suspect that attention on feature maps causes the feature maps to become artificially small, resulting in numerical instability.
In
all
cases, the performance is slightly worse than what CNe~\cite{Yi18a} gives in \Table{ablation} (.414) showing that attention on feature maps is actually harmful.

We also tried modifying the output of the network -- \ie, the weights used by the eight-point algorithm -- to use the ReLU$+$Tanh configuration from \cite{Yi18a}, which we report in the bottom row of \Table{attfeat}.
This variant trained in a stable way, but provided sub-par results that are always lower than using local attention only on the normalization.
Note that with ReLU$+$Tanh and local attention only, attention on the feature maps does help -- by 1\% relative -- but the increase in performance is very small compared to what our method can achieve.

\section{Essential matrix estimation -- \Table{essential}}
\label{sec:e_vs_f}
Several learned methods~\cite{Yi18a,Zhang19a} focus on estimating the Essential matrix instead of the Fundamental matrix.
The latter is more broadly applicable as it does not need a-priori knowledge about the intrinsics of the camera -- hence it is closer to Computer Vision ``in the wild''.

We now demonstrate that our approach outperforms these methods for Essential matrix estimation as well.
We report the results in \Table{essential}, using the authors' original implementations for this comparison.
We also report the performance of robust estimators such as RANSAC and MAGSAC. 
For RANSAC, we rely on \texttt{findEssentialMat} from OpenCV.
We found that it is beneficial to use both local and global attention when applying *SAC to the Essential matrix problem unlike the Fundamental matrix problem, and we simply threshold $\bw$ with an optimal threshold (i.e., $10^{-7}$) found on the validation set, and feed the surviving correspondences to *SAC.
We observe that RANSAC improves the performance for ACNe in the outdoor experiments.
This is due to the reduced complexity of the problem, which assumes known camera intrinsics.
For MAGSAC, we carefully implement the 5-point algorithm into their framework for estimating the Essential matrix.
While it achieves competitive results, MAGSAC is still much worse than RANSAC because it is originally geared for Fundamental matrix estimation.

\begin{table}[tbh!]
\begin{center}
\small
\setlength\tabcolsep{12pt} %
\begin{tabular}{lcc}
\toprule
 Method & {Outdoors} & {Indoors}\\
\midrule
 RANSAC &  {.671}  & {.365}  \\
 MAGSAC &  {.415}  & {.204}  \\
\midrule
 CNe (weighted-8pt) &  {.515}  & {.332}  \\
 CNe+RANSAC &  {.750}  & {.404}  \\
 CNE+MAGSAC &  {.514}  & {.259}  \\
 DFE (weighted-8pt) &  {.573}  & {.352}  \\
 DFE+RANSAC &  {.721}  & {.384}  \\
 DFE+MAGSAC &  {.532}  & {.265}  \\
 OANet (weighted-8pt) &  {.648}  & {.401}  \\
 OANet+RANSAC &  \underline{.776}  & \underline{.419}  \\
 OANet+MAGSAC &  {.547}  & {.271}  \\
\midrule
 ACNe (weighted-8pt) & {.706} & {\bf .429} \\
 ACNe+RANSAC & {\bf .780} & {.418} \\
 ACNe+MAGSAC &  {.415}  & {.187}  \\
\bottomrule
\end{tabular}
\end{center}
\caption{
{\bf Essential matrix estimation -- }
mAP at a 20$^\circ$ error threshold when we train the models to estimate the Essential matrix instead of the Fundamental matrix, for the indoors and outdoors experiments.
} %
\label{tbl:essential}
\end{table}
\begin{table}[!ht]
\begin{center}
\small
\setlength\tabcolsep{10pt} %
\begin{tabular}{lccc}
\toprule
\#ACN per ARB & 1 & 2 & 3\\
\midrule
Weighted-8pt & {.527} & {.602} & {.621}  \\   
\bottomrule
\end{tabular}
\end{center}
\caption{
{\bf Ablation on \#ACN} -- Performance as we vary the number of ACNs within each ARB -- mAP at $20^\circ$ on our validation set -- \emph{Saint Peter's Square}.
}
\label{tbl:num_acn}
\end{table}

\section{Number of ACNs within ARB -- \Table{num_acn}}
\label{sec:sec_num_acn}

We perform an ablation study to evaluate the impact of the number of ACN blocks within each ARB. 
Due to the increasing computation overhead and GPU memory limitations, we only report the results of ACNe up to three ACN blocks for each ARB; see~\Table{num_acn}.
We expect that more ACN blocks would further improve the accuracy, at the cost of an increase in memory/computation.
We use 2 blocks, as it provides a good compromise between computational requirements and performance, and also the additional advantage that this makes our results directly comparable to CNe~\cite{Yi18a}.

\section{Number of parameters -- \Table{num_weights}}
\label{sec:num_para}
Even though ACNe significantly outperforms CNe, the number of parameters added to CNe is only 6K, which is only ${\approx}1.5\%$ more.
The advantages are more prominent when we compare against OANet, which introduces a \textit{significant} increase in the number of parameters in the network, while providing worse results.

\begin{table}%
\begin{center}
\setlength\tabcolsep{8pt} %
\begin{tabular}{lccc}
\toprule
Methods & OANet & CNe& ACNe\\
\midrule
\# of Parameters & {2347K} & {394K} & {400K}  \\   
\bottomrule
\end{tabular}
\end{center}
\caption{
{\bf Number of parameters --} Our method introduces a small overhead compared to CNe, and is much smaller than OANet.
}
\label{tbl:num_weights}
\end{table}
\begin{table}%
\begin{center}
\setlength\tabcolsep{6pt} %
\begin{tabular}{lcc}
\toprule
Methods & Network & Robust estimator \\
\midrule
RANSAC & --- & 194\\   
MAGSAC & --- & 2752\\   
CNe & {15}& ---  \\   
CNe + RANSAC & {15}& {19}  \\   
CNe + MAGSAC & {15}& {523}  \\   
OANet & {18}& ---  \\   
OANet + RANSAC & {18}& {13}  \\   
OANet + MAGSAC & {18}& {546}  \\   
ACNe & {14}& ---  \\   
ACNe + RANSAC & {14}& {16}\\   
ACNe + MAGSAC & {14}& {594}\\   
\bottomrule
\end{tabular}
\end{center}
\caption{
{\bf Average elapsed time -- } Runtime, in milliseconds, for individual steps of each method. 
We execute the forward pass of our networks on a GTX 1080 Ti GPU and the robust estimator on a Intel(R) Core(TM) i7-8700 CPU.
We disable multi-threading for the CPU timings since not all robust estimator implementations support multi-threading.
Networks are implemented with TensorFlow 1.8.0, except for OANet, which uses PyTorch 1.2.0.}
\label{tbl:timing}
\end{table}

\section{Timing of *SAC Methods -- \Table{timing}}
\label{sec:timing_sac}
We observed that ACNe and CNe share a similar runtime, and are both more efficient than OANet, which uses a deep permutation-equivariant network that performs an iterative refinement of an initial guess.
Additionally, due to the GPU efficiency and low computational complexity, the runtime of learned methods is negligible compared with traditional robust estimators (*SAC).
Furthermore, learning-based methods are capable of facilitating the task of a robust estimator by proactively rejecting outliers.
For instance, we found that ACNe makes RANSAC approximately $12\times$ times and MAGSAC approximately $5\times$ faster, while also significantly improving overall performance; see~\Table{stereo}.

\balance

\begin{figure}[t!]
\includegraphics[width=1.0\linewidth, height=0.8\linewidth]{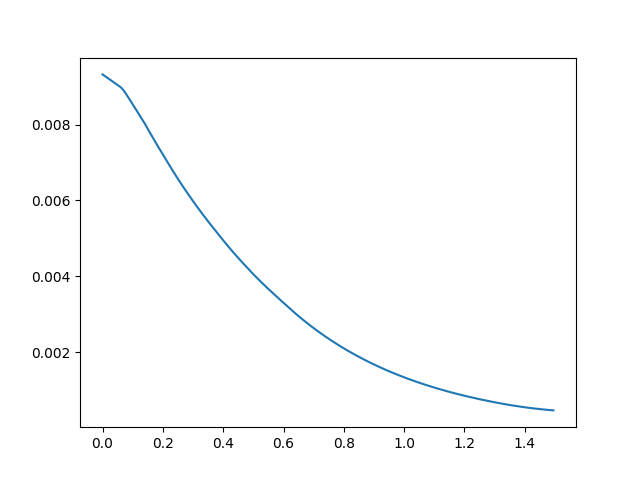}
\vspace{-1em}
\caption{
{\bf Output weight computed by the 2-layer MLP for a given residual input -- } Computed for the \emph{residual-to-weight} variant of ACNe.
}
\label{fig:r2w}
\end{figure}
\begin{table}%
\begin{center}
\setlength\tabcolsep{10pt} %
\begin{tabular}{lccc}
\toprule
Method & CNe & IRLS & ACNe \\
\midrule
$\ell_2$ error& {.0038} & {.0024} & \textbf{.0008}   \\   
\bottomrule
\end{tabular}
\end{center}
\caption{
{\bf Performance of learnt \emph{IRLS} --} Comparison with CNe and ACNe on the line fitting problem, under an outlier ratio of 0.7. 
}
\label{tbl:r2w}
\end{table}

\section{Learnt IRLS variant -- \Figure{r2w}, \Table{r2w}}
\label{sec:residual2weight}
Our method is \textit{inspired} by iteratively re-weighted least squares~(IRLS), and not a direct translation.
We learn a representations-to-weights mapping, and not a residuals-to-weights mapping as in traditional IRLS; see~\eq{irls}.
To ensure the validity of our approach, we compare to a variant of our method that is more faithful to IRLS.
In each (unrolled) iteration, we compute residuals by solving for the final objective (\eg the optimal line parameters) with the current weights.
Then, we represent $\psi(.)^{-1}$ in \eq{irls} with a 2-layer MLP that is \textit{shared} between iterations.
In other words, the MLP serves as a \textit{learnt kernel}, which is traditionally hand-picked in the IRLS literature.
We evaluate performance on the line fitting example due to its simplicity.

As shown in \Table{r2w}, the \emph{residuals-to-weights} variant performs significantly worse than ACNe.
This is not surprising, given that the IRLS variant is more restricted in what in can do, compared to ACNe.
However, it is interesting to note that it still performs better than classical CNe.
One very interesting aspect is that the learnt $\psi(.)^{-1}$ has the typical monotonically decreasing property of typical (M-Estimator) kernel functions; see \Figure{r2w}.

\end{document}